\pdfoutput=1
\documentclass[10pt,logo,copyright]{nvidiatechreport}
\usepackage[authoryear,sort&compress,round]{natbib}

\usepackage[utf8]{inputenc} % allow utf-8 input
\usepackage[T1]{fontenc}    % use 8-bit T1 fonts

\usepackage{parskip}        % no paragraph indents
\usepackage{url}            % simple URL typesetting
\usepackage{hyperref}       % hyperlinks
\usepackage{booktabs}       % professional-quality tables
\usepackage{amsfonts}       % blackboard math symbols
\usepackage{nicefrac}       % compact symbols for 1/2, etc.
\usepackage{microtype}      % microtypography
\usepackage{xcolor}         % colors
\usepackage[dvipsnames]{xcolor} % more color names
\usepackage{graphicx}
\usepackage{animate}        % for 360 video in teaser
\usepackage{subcaption}
\usepackage{tabularx}
\usepackage{makecell}
\usepackage{adjustbox}
\usepackage{setspace}
\newcolumntype{M}[1]{>{\centering\arraybackslash}m{#1}}
\usepackage{float}
\usepackage{tikz}
\usetikzlibrary{positioning,shapes,arrows}
\usepackage{amsmath,amsfonts,bm, bbm,leftindex}
\usepackage{multirow}
\usepackage{comment}
\usepackage{gensymb}
\usepackage{lipsum}
\usetikzlibrary{arrows.meta, positioning, fit}
\usepackage[para]{threeparttable}
\usepackage{tikz}
\usetikzlibrary{tikzmark}

\def\eqref#1{equation~\ref{#1}}
\def\1{\bm{1}}

\DeclareMathAlphabet{\mathsfit}{\encodingdefault}{\sfdefault}{m}{sl}
\SetMathAlphabet{\mathsfit}{bold}{\encodingdefault}{\sfdefault}{bx}{n}
\makeatletter
\let\save@mathaccent\mathaccent
\newcommand*\if@single[3]{%
  \setbox0\hbox{${\mathaccent"0362{#1}}^H$}%
  \setbox2\hbox{${\mathaccent"0362{\kern0pt#1}}^H$}%
  \ifdim\ht0=\ht2 #3\else #2\fi
  }
\newcommand*\rel@kern[1]{\kern#1\dimexpr\macc@kerna}
\newcommand*\widebar[1]{\@ifnextchar^{{\wide@bar{#1}{0}}}{\wide@bar{#1}{1}}}
\newcommand*\wide@bar[2]{\if@single{#1}{\wide@bar@{#1}{#2}{1}}{\wide@bar@{#1}{#2}{2}}}
\newcommand*\wide@bar@[3]{%
  \begingroup
  \def\mathaccent##1##2{%
    \let\mathaccent\save@mathaccent
    \if#32 \let\macc@nucleus\first@char \fi
    \setbox\z@\hbox{$\macc@style{\macc@nucleus}_{}$}%
    \setbox\tw@\hbox{$\macc@style{\macc@nucleus}{}_{}$}%
    \dimen@\wd\tw@
    \advance\dimen@-\wd\z@
    \divide\dimen@ 3
    \@tempdima\wd\tw@
    \advance\@tempdima-\scriptspace
    \divide\@tempdima 10
    \advance\dimen@-\@tempdima
    \ifdim\dimen@>\z@ \dimen@0pt\fi
    \rel@kern{0.6}\kern-\dimen@
    \if#31
      \overline{\rel@kern{-0.6}\kern\dimen@\macc@nucleus\rel@kern{0.4}\kern\dimen@}%
      \advance\dimen@0.4\dimexpr\macc@kerna
      \let\final@kern#2%
      \ifdim\dimen@<\z@ \let\final@kern1\fi
      \if\final@kern1 \kern-\dimen@\fi
    \else
      \overline{\rel@kern{-0.6}\kern\dimen@#1}%
    \fi
  }%
  \macc@depth\@ne
  \let\math@bgroup\@empty \let\math@egroup\macc@set@skewchar
  \mathsurround\z@ \frozen@everymath{\mathgroup\macc@group\relax}%
  \macc@set@skewchar\relax
  \let\mathaccentV\macc@nested@a
  \if#31
    \macc@nested@a\relax111{#1}%
  \else
    \def\gobble@till@marker##1\endmarker{}%
    \futurelet\first@char\gobble@till@marker#1\endmarker
    \ifcat\noexpand\first@char A\else
      \def\first@char{}%
    \fi
    \macc@nested@a\relax111{\first@char}%
  \fi
  \endgroup
}
\makeatother
\usepackage[nameinlink,noabbrev]{cleveref}
\crefname{equation}{Eq.}{Eqs.}
\crefname{figure}{Fig.}{Figs.}
\crefname{section}{Sec.}{Sec.}
\crefname{table}{Tab.}{Tabs.}

\newcommand{\maven}{\textsc{Maven}\xspace}
\newcommand{\msted}{\textsc{MSTED}\xspace}
\newcommand{\crbase}{CR2\xspace}

\renewcommand{\today}{}

\title{\maven: A Multi-stage Agentic Annotation Pipeline for Video Reasoning Tasks}

\author{Han Zhang \quad Wanting Jiang \quad Tomasz Kornuta \quad Tian Zheng \quad Vidya Murali\\
NVIDIA}

\keywords{video reasoning; annotation pipeline; chain-of-thought; agentic workflows; synthetic data; vision-language models}

\begin{abstract}
Training Vision Language Models (VLMs) for video event reasoning requires high-quality structured annotations capturing not only what happened, but when, where, why, and with what consequence, at a scale manual labelling cannot support.
We present \maven (Multi-stage Agentic Video Event aNnotation), a multi-stage agentic pipeline that turns raw videos into multi-task training data with Chain-of-Thought (CoT) reasoning traces, organized around a designated \emph{Event of Focus}.
At its core, \maven synthesizes a \emph{Multi-Scale Spatio-Temporal Event Description} (MSTED) from three complementary caption levels; this explicit intermediate serves as the sole input to downstream Q\&A generation across multiple task formats.
Crucially, \maven supports \emph{agent-driven domain adaptation}: given a new video dataset and target question examples, the agent redesigns all prompts top-down without manual re-engineering.
A hierarchical refinement loop further classifies annotation errors against a taxonomy, traces root causes to the originating pipeline stage, and applies targeted edits that rewrite prompts or modify the pipeline structure itself, iteratively improving data quality.
We apply \maven to label over 5,300 traffic videos and fine-tune Cosmos-Reason2-8B on the resulting data.
On a private CCTV evaluation set, fine-tuning surpasses both Gemini 2.5 Pro and 3.1 Flash, including a $+38.8$-point gain in MCQ accuracy over zero-shot.
On AccidentBench, CCTV-only training lifts Cosmos-Reason2 by $+10.7$ MCQ points and matches Gemini 2.5 Pro despite seeing no dashcam videos; adding agent-adapted dashcam annotations narrows the gap to Gemini 3.1 Flash, and RL post-training pushes overall performance past both Gemini baselines.
Qualitative results on warehouse surveillance and public safety videos further show the agentic workflow readily adapts the pipeline to new domains.
\end{abstract}

\begin{document}

\maketitle

\vspace{-2em}
\small
\textbf{Links:
\href{https://github.com/NVIDIA-TAO/tao-data-services/tree/main/nvidia_tao_ds/auto_label/video_reasoning_annotation}{Code}
\textbar 
\href{https://github.com/NVIDIA-TAO/tao-skill-bank/tree/main/skills/data/tao-generate-video-reasoning-annotations}{\ Skill}
\textbar
\href{https://huggingface.co/datasets/nvidia/PhysicalAI-Traffic-Anomaly-Reasoning}{\ Dataset}}

\vspace{-.15in}
\begin{figure}[!h]
  \centering
  \includegraphics[width=\linewidth]{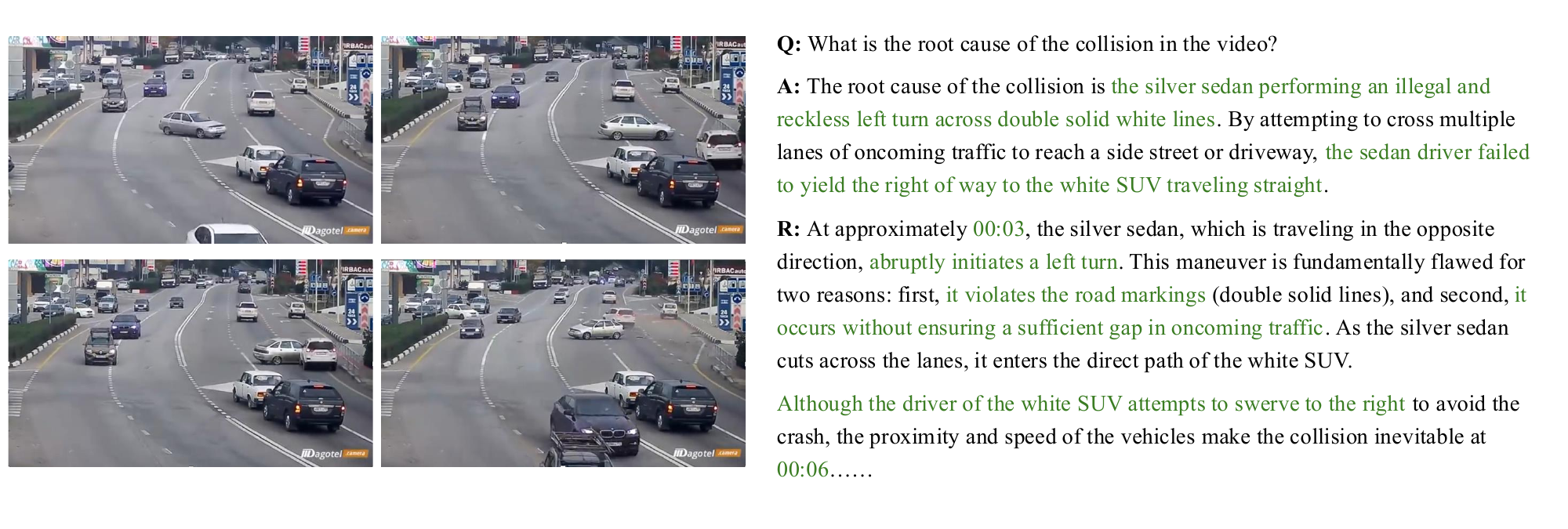}
  \vspace{-2.5em}
  \caption{\textbf{\maven-generated video reasoning annotation example.} Given a raw traffic-collision video, \maven produces a structured event representation and uses it to generate question-answer pairs with explicit reasoning traces. The example illustrates how the pipeline converts raw video evidence into training-ready supervision for video reasoning models.}
  \label{fig:teaser}
\end{figure}

\abscontent
\normalfont\normalsize

\section{Introduction}
\label{sec:intro}

Training Vision Language Models (VLMs) for video event reasoning poses a common analytical demand: understanding \emph{who} was involved, \emph{what} sequence of events unfolded, \emph{where} and \emph{when} key moments occurred, and \emph{why} the event happened.
This structured causal reasoning is critical for intelligent transportation systems, workplace safety monitoring, and physical AI, yet remains precisely where recent VLMs fall short despite strong general video understanding.

The central challenge is training data.
Structured chain-of-thought (CoT) annotations for video events (descriptions capturing temporal dynamics, spatial relationships, causal factors, and multi-step reasoning) are expensive to produce at scale.
Existing auto-labeling approaches~\citep{cosmos_reason1, vad-r1, alpamayo} either rely on flat single-pass video descriptions that permanently lose fine-grained detail, or are constrained to fixed taxonomies and domain-specific sensor inputs, limiting generalizability.
None produce a reusable intermediate event representation that can ground diverse downstream task formats from a single annotation pass.

We present \maven (Multi-stage Agentic Video Event aNnotation), a multi-stage agentic pipeline that addresses this gap.
We define the \emph{Event of Focus} (EoF) as any notable event in video, whether anomalous or routine within its domain, that the pipeline should characterize and generate training data around.
The core design principle is to construct the most complete structured representation of the scene and Event of Focus \emph{before} generating any downstream annotation.
\maven proceeds in three stages: (1)~three-level video captioning capturing global context, dense timestamped events, and fine-grained chunk-level detail; (2)~synthesis of a \emph{Multi-Scale Spatio-Temporal Event Description} (\msted) that consolidates all caption levels into a structured characterization of the Event of Focus; and (3)~generation of multi-task CoT Q\&A (MCQ, binary verification, and open-ended) grounded solely on the \msted.
Because all downstream annotations derive from the \msted rather than the raw video, the Q\&A generator cannot hallucinate details absent from the structured representation, and human reviewers can audit the \msted as a natural verification checkpoint before large-scale Q\&A generation.
Figure~\ref{fig:teaser} shows a representative traffic-collision example in which \maven converts raw video evidence into a grounded question, answer, and reasoning trace.

Crucially, \maven supports \emph{agent-driven domain adaptation}: given target benchmark question examples and a new video domain description, an agent consultation workflow redesigns prompts top-down across all pipeline stages, adapting the pipeline to new video domains, camera views, event types, and question styles automatically without manual re-engineering.

Beyond one-shot adaptation, \maven supports \emph{hierarchical pipeline refinement} from human feedback. When reviewers identify systematic annotation issues, the agent classifies errors against a structured taxonomy, traces each root cause through the pipeline hierarchy to the originating stage, and applies targeted fixes: rewriting prompts for gaps the current configuration misses, or inserting new pipeline stages for structural limitations that prompt changes alone cannot address. This distinguishes \maven from static pipelines that degrade silently when applied to challenging video distributions.

We apply \maven to label over 5,300 traffic videos (3,841 CCTV and 1,500 dashcam) and fine-tune Cosmos-Reason2~\citep{cosmos_reason2} on the resulting data.
On our private CCTV evaluation set, fine-tuning yields $+38.8$, $+35.0$, and $+24.1$ point improvements in MCQ accuracy, verification accuracy, and open-ended score over zero-shot, respectively, surpassing both Gemini~2.5~Pro and Gemini~3.1~Flash~\citep{comanici2025gemini} on all three metrics.
On the public AccidentBench~\citep{accidentbench} benchmark, our CCTV-trained model, which has never seen dashcam footage during training, matches Gemini~2.5~Pro, demonstrating that the structured CoT reasoning capability induced by the pipeline is \emph{generalizable} rather than domain-specific.
Adding agent-adapted dashcam annotations narrows the gap to Gemini~3.1~Flash, and RL post-training pushes overall performance past both Gemini baselines while CCTV evaluation set performance remains stable.
We additionally demonstrate qualitative generalization of the agentic pipeline to warehouse surveillance and public safety domains.

\noindent\textbf{Contributions:}
\begin{itemize}
    \item \textbf{Pipeline.} \maven: a multi-stage agentic pipeline producing structured \msted descriptions and multi-task CoT Q\&A from raw videos, structured around Events of Focus with an explicit intermediate representation that avoids the information loss of single-pass approaches.
    \item \textbf{Agentic domain adaptation.} An agent consultation workflow, packaged as a single Agent Skill~\citep{anthropic_skills}, that adapts the pipeline to new domains and question styles given only a domain description and example questions, requiring no manual prompt engineering.
    \item \textbf{Hierarchical refinement.} A structured three-stage refinement process (error taxonomy classification, root cause tracing through the pipeline hierarchy, and targeted configuration edits) that distinguishes prompt gaps from structural limitations and resolves each appropriately.
    \item \textbf{Dataset.} 3,841 CCTV and 1,500 dashcam traffic videos labeled with diverse CoT training data across three task formats; qualitative demonstrations on warehouse and public safety domains.
    \item \textbf{Results.} Empirical evidence that structured intermediate representations enable domain-general reasoning: CCTV-only training matches Gemini~2.5~Pro on dashcam benchmarks, and agent-adapted dashcam training with RL post-training exceeds both Gemini baselines without degrading in-domain performance.
\end{itemize}

\section{Related Work}
\label{sec:related}

\subsection{Video Anomaly Datasets and Benchmarks}
\label{sec:related-datasets}

Prior video anomaly datasets largely target frame- or clip-level classification rather than structured reasoning.
UCF-Crime~\citep{ucfcrime} established the weakly-supervised surveillance paradigm with 1{,}900 videos across 13 categories, while CADP~\citep{cadp} provides spatio-temporal annotations.
None include the causal reasoning chains (why the accident occurred, what behaviors contributed, what followed) needed to train reasoning VLMs.

Recent VLM-oriented benchmarks have begun to fill this gap.
AccidentBench~\citep{accidentbench} provides ${\sim}19{,}000$ human-annotated MCQ pairs stratified by difficulty and reasoning type, including temporal reasoning, spatial reasoning, and intent goal reasoning; its land split is our primary public benchmark.
SurveillanceVQA-589K~\citep{surveillancevqa} shows the scale achievable with AI-assisted labeling for open-ended Q\&A.
These works evaluate reasoning but do not themselves generate training data; \maven addresses this gap by producing diverse CoT Q\&A grounded in structured event representations.

\subsection{Auto-labeling and Training Data for Video VLMs}
\label{sec:related-autolabel}

General-purpose video-language models~\citep{qwen3,internvl} show strong video understanding, but structured causal reasoning (fault attribution, temporal localization, consequence prediction) remains weak without targeted fine-tuning.
The bottleneck is data: chain-of-thought annotation at scale requires either prohibitive human effort or automated pipelines that preserve fine-grained detail.
The dominant approach uses stronger models to label reasoning data for weaker, deployable ones.
% Earlier video-instruction efforts such as VideoChat~\citep{videochat} and Video-LLaMA~\citep{videollama} established video-grounded instruction tuning but do not target structured causal reasoning.
Cosmos-Reason1~\citep{cosmos_reason1} compresses each video into a single global description before synthesizing Q\&A, which is scalable but lossy.
VAD-Reasoning~\citep{vad-r1} concatenates per-frame captions and prompts an LLM for anomaly explanations, missing events between sampled frames.
Alpamayo-R1~\citep{alpamayo} targets ego-centric autonomous driving within a closed taxonomy and requires proprietary sensor metadata unavailable in general surveillance, while VLM-AutoDrive~\citep{vlm_autodrive} post-trains VLMs on dashcam safety-critical events within a fixed task formulation.
Closest to our data pipeline, LongVILA-R1~\citep{longvila_r1} chunks long videos, captions each chunk, and uses an LLM to consolidate the chunk captions into CoT training data for long-video reasoning.
Though not directly comparable, these designs inspired \maven's multi-scale captioning and its emphasis on an explicit intermediate representation.

\maven differs from these approaches structurally in two ways.
First, rather than flat chunk captions alone, \maven produces a \emph{hierarchical} three-level decomposition that is mutually corrective across scales.
Second, rather than consolidating directly into Q\&A, \maven synthesizes an explicit structured intermediate (the \msted) that serves as a verification checkpoint and the sole input to downstream Q\&A generation, avoiding irrecoverable information loss while enabling multiple task formats from a single annotation effort.

\subsection{Agentic Pipelines for Video Understanding}
\label{sec:related-agentic}

LLM-based agents are an emerging tool for both video analysis and data curation.
At inference time, several frameworks apply agentic reasoning to anomaly detection: QVAD~\citep{qvad} treats VLM-LLM interaction as a dynamic dialogue, PANDA~\citep{panda} deploys a planning-and-reflection ``AI Engineer'' for VAD, Follow the Rules~\citep{followrules} translates normality definitions into textual rules, and VERA~\citep{vera} optimizes guiding questions offline.
These improve how models reason about anomalies at test time but do not address the upstream problem of generating structured training data.

At data-generation time, Colon-Bench~\citep{colonbench} introduces a multi-stage agentic pipeline for dense colonoscopy annotation, integrating temporal proposals, tracking, AI confirmation, and human review; we share their orchestration principle but target structured reasoning annotations rather than spatial detection.
Pipeline-level prompt optimization has also been explored outside video: single-stage optimizers such as DSPy~\citep{dspy} and OPRO~\citep{opro} address individual tasks, while multi-stage optimizers like MIPRO~\citep{mipro} treat the pipeline as a black box, unable to trace errors to their originating stage or distinguish prompt deficiencies from model limitations.
\maven's agent, by contrast, performs backward inference from target tasks to derive stage-level requirements and modifies pipeline \emph{structure} (not just prompt text) based on human feedback.
Unlike inference-time agents (QVAD, PANDA), which operate on fixed models, our agent operates at \emph{pipeline design time}; unlike Colon-Bench's fixed pipeline, the \maven agent reconfigures top-down for each new domain without manual prompt engineering, making domain adaptation a first-class capability.

\section{Method}
\label{sec:method}

\subsection{The \maven Pipeline}
\label{sec:pipeline}

\begin{figure*}[]
    \centering
    \includegraphics[width=\linewidth]{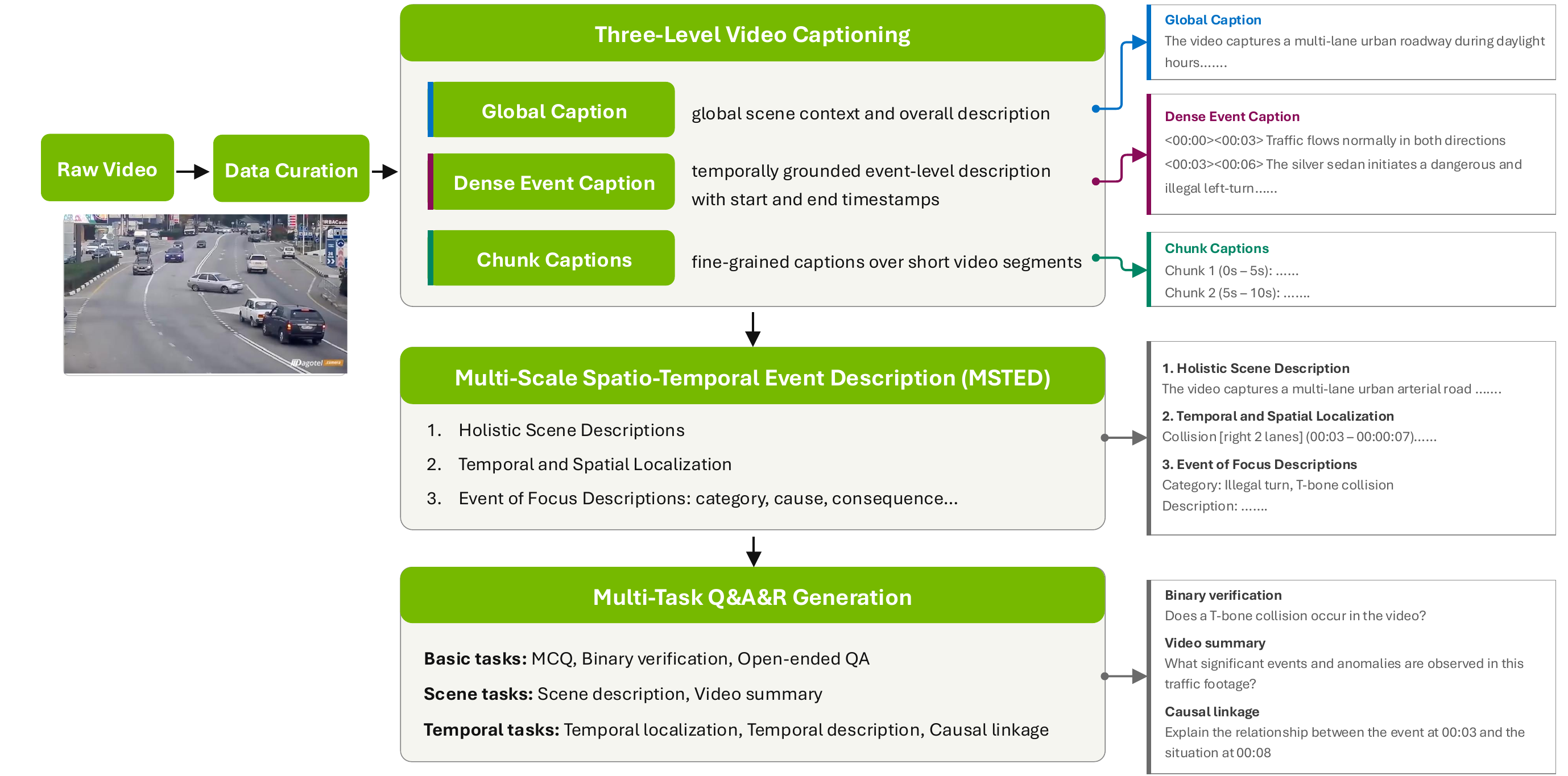}
    \vspace{-1.5em}
    \caption{\textbf{The \maven pipeline.} Raw video is converted into complementary evidence streams through global captioning, dense timestamped event captioning, and fine-grained chunk or highlight captioning. These signals are synthesized into a \msted, which serves as the sole source for downstream multi-task Q\&A generation with reasoning traces.}
    % \vspace{-0.5em}
    \label{fig:annotation_pipeline}
\end{figure*}

\maven transforms raw videos into structured CoT training data organized around \emph{Events of Focus} (EoF), defined as any notable events in the video whether anomalous or routine, through three sequential stages.
Figure~\ref{fig:annotation_pipeline} summarizes this annotation data path from raw video to \msted to multi-task reasoning supervision.
An EoF may be anomalous (\eg, a traffic accident in smart-city surveillance, or a worker safety incident in warehouse monitoring) or routine (\eg, a pedestrian crossing at a signalized intersection, or a pick-and-place operation at a workstation); what matters is that it is the salient event the pipeline characterizes and generates training data around.
For each video, the EoF is selected automatically as the most salient event surfaced during Stage~1 captioning; when a source dataset provides an explicit event label (\eg, accident class), that label is used to anchor the EoF.

The pipeline operates on a configuration (prompts and structure) produced by the top-down adaptation workflow (Section~\ref{sec:agent}). When the hierarchical refinement loop (Section~\ref{sec:refinement}) identifies a structural limitation, it can insert new stages or modify existing ones without altering the core three-stage design.
The key design principle is to construct the most complete structured representation of the scene and event \emph{before} generating any downstream annotation, factoring the challenges of video understanding, event synthesis, and task generation into distinct stages with explicit intermediate representations.

\paragraph{Stage 1: Three-level video captioning.}
A single-pass video caption cannot simultaneously capture global scene context, precise event timing, and fine-grained local detail.

We generate three complementary levels using a video VLM (\eg, Gemini~3.1~Pro):
(i)~\emph{Global caption}: a holistic description capturing scene layout, weather, time of day, and pre-event baseline behavior, establishing the context against which a notable behavior is recognized;
(ii)~\emph{Dense caption}: a temporally grounded event-level description pairing each major event with start and end timestamps, providing the causal chain and timing needed for reasoning;
(iii)~\emph{Chunk captions}: fine-grained captions over short video segments (5--30s depending on video length), recovering subtle behaviors, small objects, and brief decisive moments that a global caption of a long clip would miss.
These three levels are designed to be mutually correcting: global captions provide context that disambiguates vague chunk-level descriptions, chunk captions recover details that dense captions under-specify, and dense captions impose temporal structure that organizes chunk-level observations.

\paragraph{Stage 2: MSTED synthesis.}
An LLM (\eg, Gemini~3.1~Flash) consolidates all three caption levels into the \emph{Multi-Scale Spatio-Temporal Event Description} (\msted), consisting of three parts:
(1)~\emph{Holistic Scene Description}: environment, weather, time of day, scene layout, and pre-event baseline behavior;
(2)~\emph{Temporal and Spatial Localization}: a chronological narrative of the event's progression with precise start/end timestamps and spatial region;
(3)~\emph{Event of Focus Description}: a structured characterization of event category, temporal and spatial properties, root cause, and consequences (open-ended, with no predefined taxonomy).

The \msted serves two critical roles.
First, it acts as a \textbf{verification checkpoint}: because the \msted explicitly characterizes all salient properties of the event in a structured form, human annotators or automated validation can review it for completeness and accuracy before Q\&A generation, preventing error propagation into the training data.
Second, it is the \textbf{sole input} to Stage~3: Q\&A generators never see the raw video or original captions, only the \msted.
This factorization ensures that all generated questions are answerable from explicitly captured information; the model cannot hallucinate details absent from the structured representation.

\paragraph{Stage 3: Multi-task CoT Q\&A generation.}
A second LLM pass takes the \msted as sole context and generates three task formats, each with an explicit reasoning trace:
\textbf{MCQ} (4-option, with step-by-step reasoning on timestamps and spatial locations);
\textbf{binary verification} (yes/no questions with supporting reasoning);
\textbf{open-ended QA} (free-form questions requiring causal or descriptive reasoning).
The task types and question formats are fully configurable via prompts; no architectural changes are required.

\begin{figure*}[t]
    \centering
    \includegraphics[width=\linewidth]{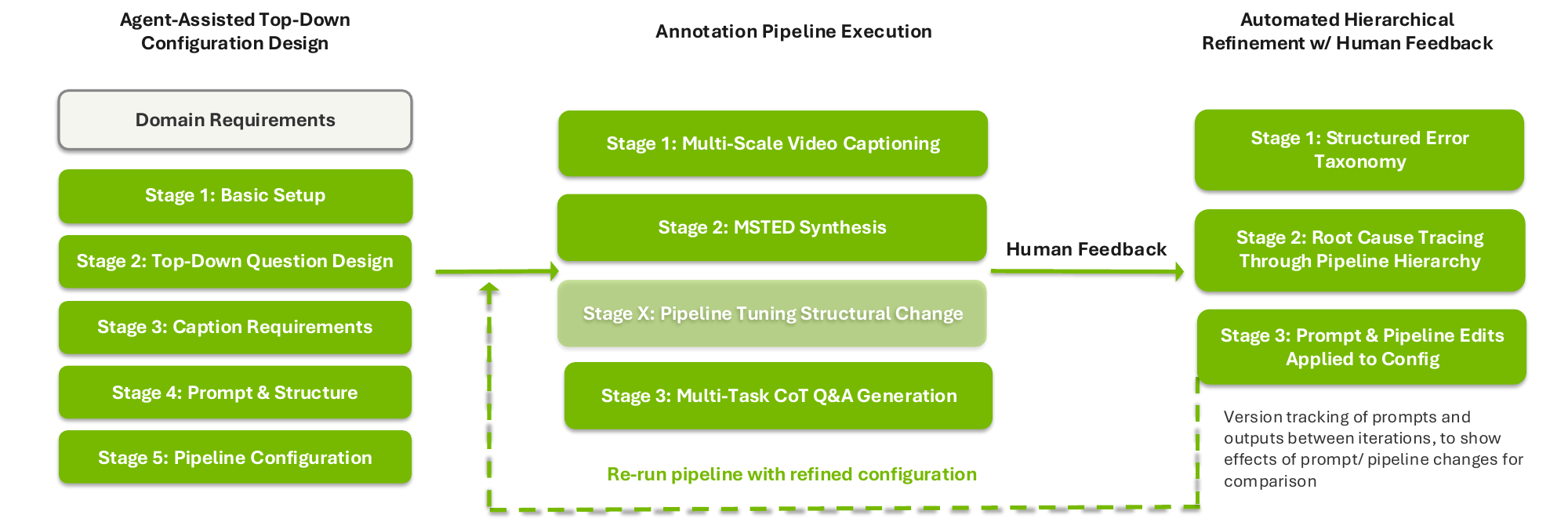}
    \vspace{-1.5em}
    \caption{The agentic approach of \maven, organized into three components: agent-assisted top-down configuration design (left), annotation pipeline execution (center), and hierarchical pipeline refinement with human feedback (right).}
    % \vspace{-1em}
    \label{fig:agentic_pipeline}
\end{figure*}

\subsection{Agent-Driven Domain Adaptation}
\label{sec:agent}

Adapting the pipeline to a new domain (\eg, from CCTV to dashcam footage) or new task types typically requires manual prompt re-engineering at each stage, a time-consuming process requiring domain expertise.
\maven automates this by packaging the pipeline as a single Agent Skill~\citep{anthropic_skills}.
As shown in Figure~\ref{fig:agentic_pipeline}, the agentic workflow wraps the annotation pipeline with top-down configuration design before execution and hierarchical refinement after human review, enabling the same core pipeline to be reconfigured for new domains and task requirements.
Following recent work~\citep{single_agent_skills_2026} showing that a multi-agent committee can be compiled into an equivalent single-agent-with-skills system at substantially lower token and latency cost, we adopt a single-orchestrator design.
The skill is instantiated via the \texttt{opencode} CLI harness backed by Claude Opus~4.6~\citep{claude-opus} with file-read, file-write, and bash tool access.
The Agent Skill abstraction is model- and harness-agnostic, so the same skill can be executed under Claude Code, Codex, or any other harness that supports Agent Skills.

The agent reads the base pipeline configuration files, the domain description, and the target question examples as context, and writes back updated prompts and any structural edits.
Given (1)~the base pipeline configuration, (2)~a target domain description, and (3)~desired question types, the agent performs \emph{backward inference}: what temporal granularity, spatial relationships, and causal depth must the \msted capture to make those questions answerable?
From this analysis, it derives a per-stage must-capture checklist, ensuring that each captioning stage produces what downstream stages will need, then rewrites all prompts to satisfy these requirements, adjusting for domain-specific visual characteristics, camera perspectives, and event semantics.

For example, when adapting from CCTV to dashcam footage targeting AccidentBench~\citep{accidentbench}, the agent adjusted captioning prompts for ego-vehicle perspective and motion dynamics, and strengthened Q\&A generation to emphasize intent attribution and temporal reasoning, question types that dominate the target benchmark.
The result is a complete domain-adapted pipeline configuration that can be applied to new video batches without further human intervention.

\subsection{Hierarchical Pipeline Refinement with Human Feedback}
\label{sec:refinement}

Beyond one-shot domain adaptation, \maven supports iterative pipeline refinement through natural-language human feedback. When reviewers identify systematic issues in the generated annotations, the agent diagnoses root causes and updates the pipeline configuration accordingly, modifying not just prompt text but pipeline \emph{structure}.
This refinement proceeds through three structured stages.

\paragraph{Stage 1: Structured Error Taxonomy.}
Given human feedback and sampled annotation outputs, the agent classifies each discrepancy against a fixed error taxonomy: \emph{misinformation}, \emph{hallucination}, \emph{missing information}, \emph{temporal error}, \emph{spatial error}, and \emph{attribution error}.
This structured classification prevents over-diagnosis: not every output error requires a pipeline change.

\paragraph{Stage 2: Root Cause Tracing Through Pipeline Hierarchy.}
For each identified error, the agent traces responsibility back through the pipeline hierarchy, from Q\&A output through the \msted to the specific captioning level where the error originates.
It classifies the root cause as either a \emph{prompt gap} (the information was capturable but the prompt failed to elicit it, fixable by prompt modification) or a \emph{system limitation} (the information cannot be recovered from the existing pipeline stages, requiring structural intervention).

\paragraph{Stage 3: Prompt \& Pipeline Edits Applied to Configuration.}
Prompt gaps are resolved by rewriting the relevant prompt; system limitations trigger a structural change, inserting a new stage or modifying an existing one.
For instance, when a reviewer noted that uniform-length chunking sometimes splits events across chunk boundaries and yields inaccurate chunk captions, Stage~2 diagnosed this as a system limitation and Stage~3 introduced an additional captioning stage supplementing the chunk captions: \emph{event-centered highlight chunks}. An LLM first identifies the key event timestamp from the existing captions, then a variable-duration segment is extracted around it for targeted re-captioning, producing more accurate descriptions of the event itself and improving \msted temporal fidelity.

% To confirm that each change has the intended effect, \maven maintains \emph{version tracking} of prompts and outputs across iterations, enabling before-and-after comparison of annotation quality following each configuration update.

\subsection{Training Dataset}
\label{sec:dataset}

\textbf{CCTV corpus.}
We apply \maven to 3,841 open-source traffic videos from roadside CCTV cameras: 808 accident videos and 3,033 normal traffic videos.
Each video is labeled with all three task formats, yielding 3,841 MCQ, 7,682 binary verification, and 3,841 open-ended QA samples with full CoT reasoning traces.

\textbf{Dashcam corpus (agent-adapted).}
Using the agent-adapted pipeline described in Section~\ref{sec:agent}, we label 1,500 dashcam collision videos from the Nexar dataset~\citep{moura2025nexar}: 750 accident videos and 750 normal videos.
The agent-redesigned prompts target AccidentBench-style questions, producing 11,200 MCQ samples with matching difficulty gradations and task type coverage.

\textbf{Evaluation sets.}
We evaluate on two held-out sets:
(i)~\emph{Private CCTV evaluation set}: 80 traffic CCTV videos sourced from YouTube with human-labeled MSTEDs and verified Q\&A, totaling 80 MCQ, 160 binary verification, and 80 open-ended samples. Focused on causal reasoning: fault attribution, root cause identification, and consequence characterization.
(ii)~\emph{AccidentBench}~\citep{accidentbench} (land split): 1{,}630 videos with 17,069 human-annotated MCQ at three difficulty levels (easy, medium, hard) and three task types (temporal, spatial, intent).
Short videos (1{,}500) are exclusively dashcam views, while medium-length (58) and long (70) videos include a mixture of dashcam and CCTV footage.
% Table~\ref{tab:dataset} summarizes video and question-sample counts across training and evaluation sets.

% \begin{table}[t]
% \centering
% \setlength{\tabcolsep}{4pt}
% \small
% \begin{tabular}{l|c|ccc}
% \toprule
%  & \textbf{Videos} & \textbf{MCQ} & \textbf{Verif.} & \textbf{Open} \\
% \midrule
% \multicolumn{5}{l}{\emph{Training}} \\
% CCTV (\maven-labeled)             & 3{,}841         & 3{,}841  & 7{,}682 & 3{,}841 \\
% Dashcam (agent-adapted)           & 1{,}500         & 11{,}200 & ---      & ---     \\
% \midrule
% \multicolumn{5}{l}{\emph{Evaluation}} \\
% Private CCTV                      & 80              & 80       & 160      & 80      \\
% AccidentBench land                & 1{,}630 & 17{,}069 & ---      & ---     \\
% \bottomrule
% \end{tabular}
% \vspace{-0.5em}
% \caption{\maven training and evaluation set statistics. Dashcam training data are MCQ-only, targeting the AccidentBench's format.}
% \vspace{-1.5em}
% \label{tab:dataset}
% \end{table}

\subsection{Training Protocol}
\label{sec:training}

We fine-tune Cosmos-Reason2-8B (\crbase)~\citep{cosmos_reason2} on \maven-labeled data following the now-standard SFT-then-RL protocol adopted in recent video and multimodal reasoning work~\citep{deepseek_r1,longvila_r1,vad-r1,cosmos_reason1,alpamayo}: supervised fine-tuning (SFT) followed by reinforcement learning (RL).

\textbf{SFT.}
We fine-tune the full model for $3$ epochs with a learning rate of $1\mathrm{e}{-5}$, global batch size $512$, using $8$ $\times$ A100 GPUs.
Video frames are sampled at rate $2$ up to $128$ frames per video.
Each training example contains the sampled video frames, the question, and the full CoT \& answer; loss is computed on the entire CoT and answer tokens so the model learns to mimic the style and depth of the generated reasoning traces before outputting a final answer across all three task formats.

\textbf{RL (DAPO).}
On top of the intermediate SFT checkpoint after 1 epoch, we apply DAPO~\citep{dapo} for $1000$ steps with learning rate $1\mathrm{e}{-6}$.
For each prompt we sample $n{=}16$, with a prompt batch of $256$ and mini-batch size $2$. 
The policy objective uses a symmetric clip range $\epsilon_\text{low}{=}\epsilon_\text{high}{=}0.2$ and KL coefficient $\beta=0.01$.
The reward is a weighted sum of (i)~format correctness ($w_\text{format} = 0.2$), enforcing the reasoning-trace schema with thinking tags, and (ii)~answer accuracy ($w_\text{acc}=1$), exact match for MCQ and binary verification.
RL is applied only to MCQ and binary verification; open-ended performance is evaluated but not RL-optimized.

\section{Experiments}
\label{sec:experiments}

\subsection{Experimental Setup}

Following the training protocol in Section~\ref{sec:training}, we report three model variants of \crbase:
\textbf{+ CCTV SFT} (SFT on CCTV data only),
\textbf{+ Dashcam SFT} (SFT on CCTV + agent-adapted dashcam data), and
\textbf{+ RL} (RL post-training on CCTV + dashcam data, initialized from the + Dashcam SFT intermediate checkpoint).
We compare these against zero-shot \crbase, Gemini~2.5~Pro, and Gemini~3.1~Flash~\cite{comanici2025gemini}, all evaluated with the same prompts.
Evaluation is conducted on the private CCTV evaluation set and the AccidentBench land split (Section~\ref{sec:dataset}).

\subsection{Private CCTV Evaluation Set}

Table~\ref{tab:its} shows results on our private CCTV evaluation set.
+ CCTV SFT yields dramatic improvements over zero-shot \crbase across all three task formats ($+38.8$ MCQ, $+35.0$ verification, $+24.1$ open-ended points) and surpasses both Gemini~2.5~Pro and Gemini~3.1~Flash on all three metrics.
The open-ended gap is particularly striking: zero-shot \crbase and Gemini~2.5~Pro both score below $16\%$ on rescaled BertScore F1, likely because their default answer style diverges from the reference answers generated by our pipeline, whereas fine-tuning on \maven data aligns the model's output distribution with the target answer format.

Adding agent-adapted dashcam data (+ Dashcam SFT) keeps CCTV performance near-stable (Verif dips by 1.25 points; MCQ and open-ended remain essentially unchanged) despite the domain shift, indicating that the pipeline-generated labels are complementary rather than conflicting.
+ RL further improves MCQ accuracy ($86.25 \to 88.75$) with slight decreases in verification and open-ended, consistent with the RL reward signal targeting only MCQ and verification tasks.
We note that the open-ended score is rescaled BertScore F1 against pipeline-generated reference answers, which rewards lexical overlap with the training distribution; it should be read as a directional signal rather than an absolute reasoning-quality measure.

\begin{table}[t]
\centering
\setlength{\tabcolsep}{5pt}
\begin{tabular}{l|l|ccc}
\toprule
\textbf{Model} & \textbf{Training} & \textbf{MCQ} & \textbf{Verif.} & \textbf{Open} \\
\midrule
Gemini~2.5~Pro           & 0-shot  & 82.50 & 76.25 & 15.60 \\
Gemini~3.1~Flash         & 0-shot  & 80.00 & 70.63 & 23.20 \\
\crbase                  & 0-shot  & 47.50 & 50.00 & 15.37 \\
\midrule
+ CCTV SFT               & SFT     & 86.25 & \textbf{85.00} & \textbf{39.45} \\
+ Dashcam SFT            & SFT     & 86.25 & 83.75 & \textbf{39.47} \\
+ RL                     & SFT+RL  & \textbf{88.75} & 81.25 & 37.29 \\
\bottomrule
\end{tabular}
% \vspace{-0.5em}
\caption{Results on the private CCTV evaluation set (80 videos). MCQ and verification are reported as accuracy (\%); open-ended as rescaled BertScore F1 (\%).}
% \vspace{-1em}
\label{tab:its}
\end{table}

\subsection{AccidentBench}

Table~\ref{tab:accidentbench} shows MCQ accuracy on AccidentBench land split across all nine video-length-by-difficulty cells, with columns grouped by video length.

\textbf{+ CCTV SFT lifts the backbone across all video lengths.}
+ CCTV SFT improves over zero-shot \crbase by $+10.7$ points overall ($29.9 \to 40.6$), with consistent gains on every length bin: Short $32.4 \to 42.4$, Medium $34.2 \to 42.4$, and Long $23.1 \to 36.9$.
This demonstrates that the training signal from structured CoT reasoning over CCTV events transfers to the Cosmos-Reason2 backbone well beyond the training distribution, including the all-dashcam Short bin.

\textbf{Domain generalization without dashcam data.}
+ CCTV SFT matches Gemini~2.5~Pro overall ($40.6\%$ vs.\ $40.3\%$) and approaches Gemini~3.1~Flash ($42.8\%$) despite seeing no dashcam videos during training.
On long videos it exceeds Gemini~2.5~Pro by $4.4$ points and approaches Gemini~3.1~Flash; on medium videos it matches Gemini~2.5~Pro and trails Gemini~3.1~Flash by only $2.0$ points.
On short videos, it still trails both Gemini baselines, reflecting the residual domain gap that CCTV-only training cannot fully cover and motivating the agent-adapted dashcam corpus next.

\textbf{Agent-adapted dashcam data closes the short-video gap.}
Adding dashcam labels generated by the agent-adapted pipeline drives the largest improvement precisely where it is needed: Short-Avg rises from $42.4$ to $47.9$, now surpassing both Gemini baselines ($46.9$ and $46.4$), with gains across all difficulty levels.
Overall accuracy rises to $42.0\%$, approaching Gemini~3.1~Flash; medium- and long-video performance remains comparable to + CCTV SFT, indicating that the dashcam labels add short-video capability without displacing the CCTV-domain reasoning already learned.

\textbf{RL further amplifies reasoning on short videos and hard questions.}
+ RL achieves the highest overall accuracy at $44.2\%$, exceeding both Gemini~2.5~Pro ($+3.9$) and Gemini~3.1~Flash ($+1.4$).
The gains concentrate on short videos and on the hard-level questions (Short-Hard: $26.1 \to 37.7$; Medium-Hard: $34.5 \to 38.3$), consistent with the RL reward targeting MCQ accuracy on questions requiring multi-step causal reasoning.

\begin{table*}[t]
\centering
\resizebox{\textwidth}{!}{%
\begin{tabular}{l|l| ccc c ccc c ccc c c}
\toprule
 &  & \multicolumn{4}{c}{\textbf{Short}} & \multicolumn{4}{c}{\textbf{Medium}} & \multicolumn{4}{c}{\textbf{Long}} & \\
\cmidrule(lr){3-6}\cmidrule(lr){7-10}\cmidrule(lr){11-14}
\textbf{Model} & \textbf{Training} & E & M & H & Avg & E & M & H & Avg & E & M & H & Avg & \textbf{Overall} \\
\midrule
Gemini~2.5~Pro       & 0-shot  & 63.0 & 42.8 & 34.8 & 46.9 & \textbf{54.7} & 33.9 & 35.8 & 41.5 & 46.0 & 32.7 & 18.7 & 32.5 & 40.3 \\
Gemini~3.1~Flash     & 0-shot  & 63.1 & 45.1 & 31.0 & 46.4 & 51.5 & \textbf{42.7} & \textbf{38.9} & 44.4 & 48.0 & \textbf{40.0} & \textbf{25.3} & \textbf{37.8} & 42.8 \\
\crbase              & 0-shot  & 43.6 & 32.4 & 21.3 & 32.4 & 40.0 & 33.0 & 29.6 & 34.2 & 30.3 & 24.3 & 14.7 & 23.1 & 29.9 \\
\midrule
+ CCTV SFT           & SFT     & 60.4 & 40.8 & 26.1 & 42.4 & 52.3 & 40.5 & 34.5 & 42.4 & 50.2 & 37.5 & 23.0 & 36.9 & 40.6 \\
+ Dashcam SFT        & SFT     & 66.6 & 46.9 & 30.2 & 47.9 & 47.7 & 42.1 & 35.0 & 41.6 & \textbf{52.6} & 35.2 & 21.3 & 36.4 & 42.0 \\
+ RL                 & SFT+RL  & \textbf{68.4} & \textbf{50.3} & \textbf{37.7} & \textbf{52.1} & 53.0 & 42.6 & 38.3 & \textbf{44.6} & 51.0 & 35.4 & 21.6 & 36.0 & \textbf{44.2} \\
\bottomrule
\end{tabular}
}
% \vspace{-0.5em}
\caption{AccidentBench land split MCQ accuracy (\%). Columns are grouped by video length (Short/Medium/Long); nested columns report difficulty (E/M/H = Easy/Medium/Hard). Short videos are exclusively dashcam; medium and long videos mix dashcam and CCTV footage.}
% \vspace{-0.5em}
\label{tab:accidentbench}
\end{table*}

\begin{figure*}[t]
    \centering
    \includegraphics[width=\linewidth]{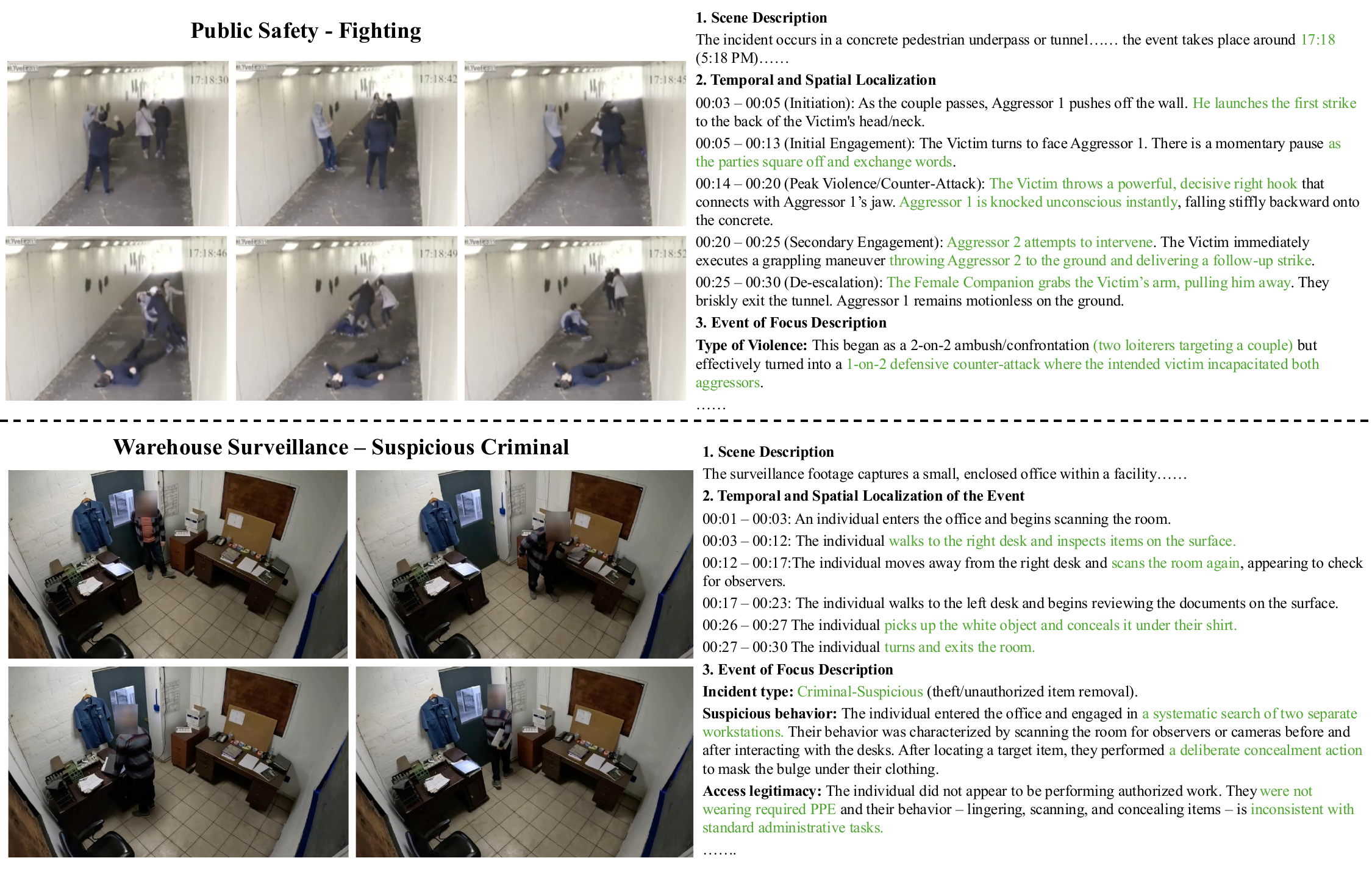}
    \vspace{-1em}
    \caption{Qualitative \maven outputs on two generalized domains. \textbf{Top (Public Safety):} a complex altercation in a pedestrian underpass, where an ambush turns into a defensive counter-attack and the initial aggressor is knocked unconscious by the intended victim. \textbf{Bottom (Warehouse Surveillance):} a suspicious-criminal case where an individual systematically searches two workstations, conceals a white object under their shirt, and exits. Highlighted spans in the \msted mark the key information that the agent-adapted pipeline captures.}
    \label{fig:qualitative}
    % \vspace{-0.5em}
\end{figure*}

\subsection{Cross-Domain Pipeline Generalization}
\label{sec:crossdomain}

To demonstrate that the agentic consultation workflow generalizes beyond traffic, we apply \maven to two additional domains by providing only a new domain description to the agent.
\emph{Public safety}: crowded-scene footage of behavioral anomalies (\eg, fights, crowd surges); the agent adjusts for dense multi-person tracking, role attribution, and intent-level reasoning.
\emph{Warehouse surveillance}: indoor CCTV of worker safety and security anomalies (\eg, falls, unsafe zone violations, unauthorized access, theft); the agent redesigns prompts for overhead camera perspective, occlusion from shelving, and safety- and security-rule-grounded behavior analysis.
In both cases the pipeline produces structured \msted outputs and Q\&A samples without manual prompt engineering or domain expertise.

Figure~\ref{fig:qualitative} shows two representative outputs.
The \emph{public safety} example captures a complex altercation in a pedestrian underpass: two loiterers ambush a passing couple, but the intended victim counter-attacks and knocks the initial aggressor unconscious, turning a 2-on-2 ambush into a 1-on-2 defensive counter-attack.
This is a challenging multi-actor temporal reasoning case, requiring the pipeline to recover both the role reversal and the causal chain connecting the successive actions, all of which the agent-adapted \msted correctly identifies.
The \emph{warehouse surveillance} example captures a suspicious criminal case: an individual enters an enclosed office, systematically searches two separate workstations, scans the room for observers, picks up a white object and conceals it under their shirt, then exits.
The \msted captures the suspicious-behavior pattern, the deliberate concealment action, and the access-legitimacy reasoning, illustrating that the agent-adapted pipeline recovers fine-grained intent attribution from multiple short actions without any manual prompt engineering.

\subsection{Ablations}
\label{sec:ablations}

\textbf{Does the structured intermediate representation matter?}
To isolate the contribution of three-level captioning and \msted synthesis, we compare \maven against a \emph{flat} baseline that generates CoT Q\&A directly from a single global caption of the same 3{,}841 CCTV videos, without the dense/chunk caption decomposition and without the \msted intermediate representation.
All other training details are held constant.

Table~\ref{tab:ablation_msted} reports the comparison.
\maven outperforms the single-pass captioning baseline on all three task formats, with gains of $+6.25$ MCQ, $+11.25$ verification, and $+3.50$ open-ended points.
The single-pass baseline itself sits at roughly the Gemini baseline level on MCQ and verification (Table~\ref{tab:its}), which indicates that a standard caption-then-generate recipe effectively distills the teachers' zero-shot performance.
\maven's three-level captioning and \msted capture and organize event information that a single global caption cannot convey; the resulting gain over the single-pass baseline is attributable to pipeline structure rather than to a stronger generator, and allows the 8B student to exceed the Gemini baselines themselves.

\begin{table}[t]
\centering
\setlength{\tabcolsep}{5pt}
\begin{tabular}{lccc}
\toprule
\textbf{Training CoT generation} & \textbf{MCQ} & \textbf{Verif.} & \textbf{Open} \\
\midrule
Single-pass Captioning          & 80.00 & 73.75 & 35.95 \\
\maven     & \textbf{86.25} & \textbf{85.00} & \textbf{39.45} \\
\bottomrule
\end{tabular}
% \vspace{-0.5em}
\caption{Ablation on the CCTV evaluation set: flat single-pass captioning vs. \maven (three-level captioning + \msted). Both variants trained on the same CCTV videos with identical SFT setup.}
% \vspace{-1.5em}
\label{tab:ablation_msted}
\end{table}

\subsection{Discussion}

\textbf{On distillation and baseline comparison.}
Our pipeline uses Gemini~3.1~Pro as the VLM for three-level captioning and Gemini~3.1~Flash as the LLM for \msted synthesis and CoT Q\&A generation, so on the surface the training labels are distilled from Gemini-class models.
A natural concern is that the fine-tuned Cosmos-Reason2-8B model simply inherits its teacher's behavior.
The ablation in Section~\ref{sec:ablations} addresses this at the data-generation level: the same teacher models combined with a single-pass pipeline fall well short of \maven, localizing the gain to pipeline structure rather than to the teacher.
Evaluation-time evidence further shows that the + RL variant reaches $44.2\%$ overall on AccidentBench, exceeding Gemini~3.1~Flash on an 8B backbone; on the private CCTV evaluation set it also outperforms both Gemini baselines across all three tasks.
Therefore, the \msted structured representation together with DAPO post-training extracts a reasoning signal from the generated data that goes beyond surface-level imitation of the generator.

\textbf{Structured representations enable cross-domain transfer.}
Surprisingly, CCTV-only training matches Gemini~2.5~Pro and approaches Gemini~3.1~Flash on AccidentBench, a dashcam-centric benchmark.
We attribute this to the \msted intermediate representation: by forcing the model to reason over structured event characterizations (temporal bounds, spatial locations, causal factors) rather than raw visual features, the training signal captures reasoning patterns that are \emph{domain-invariant}.
% A rear-end collision has the same causal structure whether viewed from a roadside camera or a dashcam.

\textbf{Agentic adaptation amplifies generalization.}
The agent-redesigned prompts for the dashcam corpus target AccidentBench question style and difficulty gradations directly, producing training data whose distribution better matches the evaluation benchmark.
The progressive improvements from + CCTV SFT to + Dashcam SFT to + RL demonstrate that data quality (from agent prompt design) and post-training methodology contribute additively, ultimately surpassing both Gemini baselines while maintaining CCTV evaluation set accuracy.

\textbf{RL disproportionately benefits hard reasoning.}
The progressive improvement from + CCTV SFT to + RL is not uniform across difficulty levels: Easy improves by $+3.1$ points, Medium by $+3.2$, and Hard by $+4.6$.
RL post-training with answer-accuracy rewards particularly strengthens the model on questions requiring multi-step causal inference (root cause identification, intent attribution, and counterfactual reasoning), which dominate the Hard split.

\section{Conclusion}
\label{sec:conclusion}

We presented \maven, a multi-stage agentic pipeline that transforms raw videos into structured CoT training data organized around a designated \emph{Event of Focus}.
By synthesizing three complementary caption levels into an explicit \msted intermediate before generating any Q\&A, \maven avoids the irrecoverable information loss of single-pass auto-labeling.
An agentic consultation workflow and a hierarchical refinement loop together enable top-down domain adaptation and error-driven pipeline evolution without manual re-engineering.

Fine-tuning Cosmos-Reason2-8B on \maven-labeled CCTV data yields a $+38.8$ MCQ-point gain on our private evaluation set and matches Gemini~2.5~Pro on AccidentBench despite seeing no dashcam videos, indicating that the induced reasoning capability is generalizable rather than domain-specific.
Adding agent-adapted dashcam labels and DAPO post-training pushes the model past both Gemini baselines overall, while CCTV performance remains stable.
Qualitative results on warehouse and public safety domains further show that the agentic workflow readily adapts the pipeline given only a domain description.

\paragraph{Limitations and future work.}
\maven currently relies on Gemini-class models for captioning and synthesis, and training gains are validated only on Cosmos-Reason2-8B; a natural next step is cross-backbone validation on other open video-language models (\eg, Qwen-VL~\citep{qwen3} and Nemotron series~\citep{nemotron}) to confirm that the observed gains transfer beyond a single model.
Other planned work includes data-scale ablations on the number of generated questions per video, component-level ablations isolating the contribution of \msted synthesis, top-down configuration, and hierarchical refinement with quantitative results, as well as cross-domain benchmarks for warehouse and public safety.

At present, \msted quality is assessed via spot-checking, and each domain adaptation takes 2--4 rounds of human review over 10--20 samples (roughly 1--2 hours).
Human reviewers excel at the detailed diagnosis that automated metrics cannot surface, but this dependence limits the scale at which the refinement loop can operate.
Our longer-horizon goal is \emph{fully automated closed-loop self-improvement}: using downstream evaluation signals to drive hierarchical refinement without human input.
Such a system would automatically attribute errors to their originating pipeline stage, distinguish prompt gaps from system limitations, and apply targeted prompt or structural fixes, progressing toward a truly \emph{self-optimizing} pipeline.

\setcitestyle{numbers}
\bibliographystyle{plainnat}
\bibliography{main}

\end{document}